# Similarity, Cardinality and Entropy for Bipolar Fuzzy Set in the Framework of Penta-valued Representation

Vasile Patrascu

Department of Informatics Technology, Tarom Company
Bucharest, Romania
Email: patrascu.v@gmail.com

*Abstract* — In this paper one presents new similarity, cardinality and entropy measures for bipolar fuzzy set and for its particular forms like intuitionistic, paraconsistent and fuzzy set. All these are constructed in the framework of multi-valued representations and are based on a penta-valued logic that uses the following logical values: true, false, unknown, contradictory and ambiguous. Also a new distance for bounded real interval was defined.

## 1 Introduction

Similarity measures play an important role in different research topics such as image analysis, pattern recognition, decision making and market prediction. In the same way, distance measure is an important tool which describes differences between two objects and considered as a dual concept of similarity measure [12]. The choice of a similarity measure or a distance measure for any fields of research is not trivial [8,10]. Since Zadeh proposed fuzzy sets [11], many scholars have conducted research on similarity measures between fuzzy sets [9,12]. Other similarity measures are proposed for Atanassov intuitionistic fuzzy sets as a generalization of fuzzy sets [2, 3, 7].

In this paper, one presents a new measure of distance for the interval $[a,b]$ and then for bipolar fuzzy values. Based on the new distance, the similarity of bipolar fuzzy values was defined. Then, using the similarity or dissimilarity, the cardinality and entropy measures are constructed for bipolar fuzzy set. All these measures are done for bipolar fuzzy values or bipolar fuzzy set in the framework of penta-valued representation [6].

The paper has the following structure: section 2 presents the fuzzy set and its extensions: intuitionistic fuzzy set, paraconsistent fuzzy set and bipolar fuzzy set. Also, the main operators for bipolar fuzzy sets are presented. Section 3 presents the penta-valued representation of bipolar fuzzy sets defining indexes of truth, falsity, unknowingness, contradiction and ambiguity. Section 4 presents a new distance measure for the interval $[a,b]$ and its particular forms for $[0,1]$ and $[-1,1]$. Section 5 presents distance and similarity measures for bipolar fuzzy sets while section 6 presents the measures for cardinality and entropy. Finally, the conclusions are presented in section 7.

## 2 The fuzzy sets and its extensions

Let $X$ be a crisp set (the space of points), with a generic element of $X$ denoted by $x$. In the framework of Zadeh theory [11], a *fuzzy set* $A$ in $X$ is characterized by the membership function $\mu: X \to [0,1]$. The non-membership function $\nu: X \to [0,1]$ is obtained by negation and thus both functions define a partition of unity, namely:

$$\mu + \nu = 1 \qquad (2.1)$$

Atanassov has extended the fuzzy sets to the *intuitionistic fuzzy sets* [1]. Atanassov has relaxed the condition (2.1) to the following inequality:

$$\mu + \nu \leq 1 \qquad (2.2)$$

He has used the third function, the index of uncertainty $\pi$ that verifies the equality:

$$\pi = 1 - \mu - \nu \qquad (2.3)$$

In the same way, we can consider instead of (2.1) the following condition:

$$\mu + \nu \geq 1 \qquad (2.4)$$

Thus, we obtain the *paraconsistent fuzzy set* [6] and one can define the index of contradiction:

$$\kappa = \mu + \nu - 1 \qquad (2.5)$$

There is a duality between intuitionistic fuzzy set and paraconsistent fuzzy sets. More generally, in this paper, we will consider as *bipolar fuzzy set* (BFS) a set $A$, defined by two functions totally independent $\mu: X \to [0,1]$ and $\nu: X \to [0,1]$. For this kind of sets, one defines the union, the intersection, the complement, the dual and the negation operators.

*The union* $A \cup B$ for two sets $A, B \in BFS$ is defined by formulae:

$$\begin{cases} \mu_{A \cup B} = \mu_A \vee \mu_B \\ \nu_{A \cup B} = \nu_A \vee \nu_B \end{cases} \qquad (2.6)$$

*The intersection* $A \cap B$ between two sets $A, B \in BFS$ is defined by the formulae:

$$\begin{cases} \mu_{A \cap B} = \mu_A \wedge \mu_B \\ \nu_{A \cap B} = \nu_A \vee \nu_B \end{cases} \qquad (2.7)$$

In formulae (2.6) and (2.7), the symbols "$\vee$" and "$\wedge$" represent any couple of t-conorm, t-norm.

*The complement* $A^c$ for the set $A \in BFS$ is defined by the formulae:

$$\begin{cases} \mu_{A^c} = \nu_A \\ \nu_{A^c} = \mu_A \end{cases} \quad (2.8)$$

The *dual* $A^d$ for the set $A \in BFS$ is defined by the formulae:

$$\begin{cases} \mu_{A^d} = 1 - \nu_A \\ \nu_{A^d} = 1 - \mu_A \end{cases} \quad (2.9)$$

The *negation* $A^n$ for the set $A \in BFS$ is defined by the formulae:

$$\begin{cases} \mu_{A^n} = 1 - \mu_A \\ \nu_{A^n} = 1 - \nu_A \end{cases} \quad (2.10)$$

If $A$ is intuitionistic fuzzy set then $A^d$ and $A^n$ are paraconsistent fuzzy sets and if $A$ is paraconsistent fuzzy set then $A^d$ and $A^n$ are intuitionistic fuzzy sets. If $A$ is a fuzzy set then it is identical with $A^d$ while $A^n$ is identical with $A^c$. The *union* and *intersection* have the following properties:

$$(A \cup B)^n = A^n \cap B^n$$
$$(A \cup B)^c = A^c \cap B^c$$
$$(A \cap B)^n = A^n \cup B^n$$
$$(A \cap B)^c = A^c \cup B^c$$
$$(A \cup B)^d = A^d \cup B^d$$
$$(A \cap B)^d = A^d \cap B^d$$

## 3 Penta-valued representations

In the paper [6] a penta-valued logic was defined using the following values: true, false, unknown, contradictory and ambiguous. Based on this logic, penta-valued fuzzy sets FP5 was constructed [6]. The FP5 sets are described by five functions: the membership function, the non-membership function, the degree of unknownness, the degree of contradiction and the degree of ambiguity. The bipolar fuzzy set (BFS) can be translated into FP5. Particular forms of bipolar fuzzy sets can also be translated to FP5, for example fuzzy sets (FS), intuitionistic fuzzy sets (IFS) and paraconsistent fuzzy sets (PFS) [6].

### 3.1 Bipolar fuzzy set as FP5

One considers the bipolar fuzzy set $A \in BFS$ defined by the membership function $\mu$ and the non-membership function $\nu$. We will define the following indexes:

The *index of truth*:
$$t = \mu \bullet \overline{\nu} \quad (3.1.1)$$

The *index of falsity*:
$$f = \overline{\mu} \bullet \nu \quad (3.1.2)$$

The *index of unknownness*:
$$u = \overline{\mu} \bullet \overline{\nu} \quad (3.1.3)$$

The *index of contradiction*:
$$c = \mu \bullet \nu \quad (3.1.4)$$

The *index of ambiguity*:

$$i = \overline{t} \bullet \overline{f} \bullet \overline{u} \bullet \overline{c} \quad (3.1.5)$$

where "$\bullet$" is the Lukasiewicz t-norm, namely
$$x \bullet y = \max(x + y - 1, 0) \quad (3.1.6)$$

and $\overline{x}$ represents the classical negation of $x$, explicitly
$$\overline{x} = 1 - x \quad (3.1.7)$$

Taking into account the formulae (3.1.6) and (3.1.7) it results the following equivalent formulae:

$$\begin{cases} t = (\mu - \nu)_+ \\ f = (\nu - \mu)_+ \\ c = (\mu + \nu - 1)_+ \\ u = (1 - \mu - \nu)_+ \\ i = 1 - |\mu - \nu| - |\mu + \nu - 1| \end{cases} \quad (3.1.8)$$

where $a_+$ and $a_-$ represent the positive and negative part of $a$, namely: $a_+ = \max(a, 0)$, $a_- = \max(-a, 0)$. These five indexes define a partition of unity, namely:
$$t + f + u + c + i = 1$$

We must remark that there exist the following two relations:
$$\begin{cases} t \cdot f = 0 \\ u \cdot c = 0 \end{cases} \quad (3.1.9)$$

In the framework of this penta-valued representation of bipolar fuzzy set, we can not have in the same time, true and false and also, we can not have in the same time, unknown and contradictory. From (3.1.8) it results the inverse transform:

$$\begin{cases} \mu = t + c + \dfrac{i}{2} \\ \nu = f + c + \dfrac{i}{2} \end{cases} \quad (3.1.10)$$

### 3.2 Fuzzy set as FP5

We consider the fuzzy set $A \in FS$ defined by the membership function $\mu$. Using formulae (3.1.8) one define the indexes of truth, falsity and ambiguity.

$$\begin{cases} t = (2 \cdot \mu - 1)_+ \\ f = (1 - 2 \cdot \mu)_+ \\ i = 1 - |2 \cdot \mu - 1| \end{cases} \quad (3.2.1)$$

The indexes of unknownness and contradiction are zero. Finally, due to the particularity (2.1) of fuzzy sets, the penta-valued representation is reduced to a three-valued one:
$$t + f + i = 1 \quad (3.2.2)$$

From (3.1.10) one obtains the inverse transform.

### 3.3 Intuitionistic fuzzy set as FP5

We consider the intuitionistic fuzzy set $A \in IFS$ defined by the membership function $\mu$ and the non-membership function $\nu$. We will translate to a penta-valued fuzzy set using formulae (3.1.8). In consequence, one defines the indexes of truth, falsity, unknowingness and ambiguity.

$$\begin{cases} t = (\mu - \nu)_+ \\ f = (\nu - \mu)_+ \\ u = 1 - \mu - \nu \\ i = \mu + \nu - |\mu - \nu| \end{cases} \quad (3.3.1)$$

The index of contradiction is zero. Finally, due to the particularity (2.2) of intuitionistic fuzzy sets, the penta-valued representation is reduced to a tetra-valued one:
$$t + f + u + i = 1$$
From (3.1.10) one obtains the inverse transform.

### 3.4 Paraconsistent fuzzy set as FP5

We consider the paraconsistent fuzzy set $A \in PFS$ defined by the membership function $\mu$ and the non-membership function $\nu$. We will translate to a penta-valued fuzzy set using formulae (3.1.8). Hence, one defines the indexes of truth, falsity, contradiction and ambiguity.

$$\begin{cases} t = (\mu - \nu)_+ \\ f = (\nu - \mu)_+ \\ c = \mu + \nu - 1 \\ i = 2 - |\mu - \nu| - \mu - \nu \end{cases} \quad (3.4.1)$$

The index of unknownness is zero. Therefore, a bivalent knowledge representation was transformed into a tetravalent one, due to the particularity (2.4) of paraconsistent fuzzy sets. The four defined indexes verify the partition of unity condition: $t + f + c + i = 1$
From (3.1.10) it results the inverse transform.

## 4 New distance on the interval [a,b]

There are important the distances defined on the interval $[0,1]$ and that is because the membership functions are defined on the $[0,1]$. The frequently formula used for the distance on the interval $[0,1]$ is the following:
$$d(x,y) = |x - y| \quad (4.1)$$
The distance (4.1) is shift invariant. Thus, the distances between the following pairs of values $(0;0.2)$, $(0.4;0.6)$ and $(0.8;1)$ are equal with 0.2. In our opinion, the distance between $(0;0.2)$ must be less then the distance between $(0.4;0.6)$. Thus, after defuzzificaton, the first pair becomes $(0;0)$ and the second becomes $(0;1)$. For the $(0;0)$, both elements will not belong while for $(0;1)$, the first element will not belong and the second will belong to the resulted crisp set. In other words, the elements described by the first pair must be more similar then the elements described by the second pair. Using (4.1), it will not mark this result. Because of that, we will propose the use of the following distance for the interval $[0,1]$:
$$d(x,y) = \frac{|x-y|}{0.5 + |x - 0.5| \vee |y - 0.5|} \quad (4.2)$$
where "$\vee$" represents the maximum function, explicitly:
$$a \vee b = \max(a,b) \quad (4.3)$$
More generally, for the interval $[a,b]$, the distance has the following parameterized form:

$$d(x,y) = \frac{|x-y|}{\frac{b-a}{2} + \left|x - \frac{a+b}{2}\right| \vee \left|y - \frac{a+b}{2}\right|} \quad (4.4)$$

For the interval $[-1,1]$ the distance formula becomes:
$$d(x,y) = \frac{|x-y|}{1 + |x| \vee |y|} \quad (4.5)$$
The function defined by (4.4) verifies the metric properties:
- $d(x,y) = 0 \iff x = y$
- $d(x,y) = d(y,x)$
- $d(x,y) + d(y,z) \geq d(x,z)$

The first two properties are evident and for the third, we will show the proof for the case (4.5). We will analyze six possibilities. In the first four, we have $y \in \{\min(x,y,z), \max(x,y,z)\}$ and for the last two, we have $y = median(x,y,z)$.

a) $x \geq z \geq y \implies d(x,y) \geq d(x,z)$ and it results
$$d(x,y) + d(y,z) \geq d(x,z)$$
b) $y \geq z \geq x \implies d(x,y) \geq d(x,z)$ and it results
$$d(x,y) + d(y,z) \geq d(x,z)$$
c) $y \geq x \geq z \implies d(y,z) \geq d(x,z)$ and it results
$$d(x,y) + d(y,z) \geq d(x,z)$$
d) $z \geq x \geq y \implies d(z,y) \geq d(x,z)$ and it results
$$d(x,y) + d(y,z) \geq d(x,z)$$
e) $x \geq y \geq z$

It results
$$\begin{cases} |y| \leq |x| \vee |z| \\ |x| \leq |x| \vee |z| \\ |z| \leq |x| \vee |z| \end{cases}$$
and
$$\begin{cases} |x| \vee |y| \leq |x| \vee |z| \\ |y| \vee |z| \leq |x| \vee |z| \end{cases}$$
It results:
$$\begin{cases} \dfrac{|x-y|}{1+|x| \vee |y|} \geq \dfrac{|x-y|}{1+|x| \vee |z|} \\ \dfrac{|y-z|}{1+|y| \vee |z|} \geq \dfrac{|y-z|}{1+|x| \vee |z|} \end{cases}$$
and summing up, one obtains:
$$d(x,y) + d(y,z) \geq \frac{|x-y|+|y-z|}{1+|x| \vee |z|} \geq \frac{|x-z|}{1+|x| \vee |z|}$$
and finally
$$d(x,y) + d(y,z) \geq d(x,z)$$
f) $z \geq y \geq x$ This case can be proven similarly to case *e)*.

## 5 Distances and similarities for bipolar fuzzy sets

Before the presentation of the distances, we must remark that we can change the bipolar representation $(\mu,\nu)$ with the following two functions:
The *degree of truth* $\tau: X \to [-1,1]$,
$$\tau = t - f \quad (5.1)$$

The *degree of neutrality* $\omega : X \to [-1,1]$,
$$\omega = c - u \quad (5.2)$$
These two functions verify the following inequality:
$$|\tau| + |\omega| \leq 1 \quad (5.3)$$
The two components $\tau$ and $\omega$ have *polarity* because can be positive or negative.

## 5.1 Distances for bipolar fuzzy set

One considers the bipolar fuzzy set $A \in BFS$ defined by the membership function $\mu$ and the non-membership function $\nu$. Let $x_1$ and $x_2$ be two bipolar fuzzy values. Using (4.5) one defines two partial distances:

$$d_\tau(x_1, x_2) = \frac{|\tau_1 - \tau_2|}{1 + |\tau_1| \vee |\tau_2|} \quad (5.1.1)$$

$$d_\omega(x_1, x_2) = \frac{|\omega_1 - \omega_2|}{1 + |\omega_1| \vee |\omega_2|} \quad (5.1.2)$$

having the following equivalent forms:

$$d_\tau(x_1, x_2) = \frac{|t_1 - t_2| + |f_1 - f_2|}{1 + |t_1| \vee |t_2| \vee |f_1| \vee |f_2|}$$

$$d_\omega(x_1, x_2) = \frac{|c_1 - c_2| + |u_1 - u_2|}{1 + |c_1| \vee |c_2| \vee |u_1| \vee |u_2|}$$

We will combine $d_\tau$ with $d_\omega$ in order to obtain a pseudo-Hamming, a pseudo-Euclidian and a pseudo-Probabilistic distance for bipolar fuzzy set:

$$d_{PH}(x_1, x_2) = \frac{|\tau_1 - \tau_2| + |\omega_1 - \omega_2|}{1 + |\tau_1| \vee |\tau_2| + |\omega_1| \vee |\omega_2|} \quad (5.1.3)$$

$$d_{PE}^2(x_1, x_2) = \frac{(\tau_1 - \tau_2)^2}{(1 + |\tau_1| \vee |\tau_2|)^2} + \frac{(\omega_1 - \omega_2)^2}{(1 + |\omega_1| \vee |\omega_2|)^2} \quad (5.1.4)$$

$$d_{PP}(x_1, x_2) = \frac{|\tau_1 - \tau_2|}{1 + |\tau_1| \vee |\tau_2|} \oplus \frac{|\omega_1 - \omega_2|}{1 + |\omega_1| \vee |\omega_2|} \quad (5.1.5)$$

where $\oplus$ represent the probabilistic sum, namely:
$$a \oplus b = a + b - a \cdot b$$

We can particularize the formulae (5.1.3), (5.1.4) and (5.1.5) for intuitionistic and paraconsistent fuzzy sets. For fuzzy sets, all these three distances are identical and one obtains the distance define for the interval [0,1] by (4.2), explicitly:

$$d_P(x_1, x_2) = \frac{2 \cdot |\mu_1 - \mu_2|}{1 + |2 \cdot \mu_1 - 1| \vee |2 \cdot \mu_2 - 1|} \quad (5.1.6)$$

In the end, we consider the following five points: *true* $T = (1,0)$, *false* $F = (0,1)$, *unknown* $U = (0,0)$, *contradictory* $C = (1,1)$ and *ambiguous* $I = (0.5, 0.5)$. All distances $d_{PH}, d_{PE}$ and $d_{PP}$ take values in the interval [0,1] and verify the following conditions:

d1) $d(x_1, x_2) = 0 \Leftrightarrow x_1 = x_2$
d2) $d(x_1, x_2) = d(x_2, x_1)$
d3) $d(T, F) = d(U, C) = 1$
d4) $d(T, I) = d(F, I) = d(C, I) = d(U, I) = 0.5$
d5) $d(x_1, x_2) = d(x_1^c, x_2^c) = d(x_1^d, x_2^d) = d(x_1^n, x_2^n)$
d6) $d(x_1, x_2)$ is monotonically increasing with $t_1 \vee t_2$, $f_1 \vee f_2$, $u_1 \vee u_2$, $c_1 \vee c_2$ and monotonically decreasing with $t_1 \wedge t_2$, $f_1 \wedge f_2$, $u_1 \wedge u_2$, $c_1 \wedge c_2$

where "$\wedge$" represents the minimum function, explicitly:
$$a \wedge b = \min(a, b) \quad (5.1.7)$$

## 5.2 Similarities for bipolar fuzzy sets

Typically, similarity between two elements is computed by negation of distance. In this paper we will use the classical negation, namely:
$$s(x_1, x_2) = 1 - d(x_1, x_2) \quad (5.2.1)$$

We will represent with $s_{PH}$, $s_{PE}$, and $s_{PP}$ the similarities obtained for distances $d_{PH}, d_{PE}$ and $d_{PP}$. The similarities $s_{PH}$, $s_{PE}$ and $s_{PP}$ take values in the interval [0,1] and verify the following conditions:

s1) $s(x_1, x_2) = 1 \Leftrightarrow x_1 = x_2$
s2) $s(x_1, x_2) = s(x_2, x_1)$
s3) $s(T, F) = s(U, C) = 0$
s4) $s(T, I) = s(F, I) = s(C, I) = s(U, I) = 0.5$
s5) $s(x_1, x_2) = s(x_1^c, x_2^c) = s(x_1^d, x_2^d) = s(x_1^n, x_2^n)$
s6) $s(x_1, x_2)$ is monotonically decreasing with $t_1 \vee t_2$, $f_1 \vee f_2$, $u_1 \vee u_2$, $c_1 \vee c_2$ and monotonically increasing with $t_1 \wedge t_2$, $f_1 \wedge f_2$, $u_1 \wedge u_2$, $c_1 \wedge c_2$.

where "$\vee$" represents the maximum function (4.3) and "$\wedge$" represents the minimum function (5.1.7).

Next, we will consider some particular pairs with bipolar fuzzy values and we will compute their similarities. We will consider the following pairs:
$P1 = \{(0.8; 0.2), (1; 0)\}$, $P2 = \{(0.6; 0.4), (0.4; 0.6)\}$,
$P3 = \{(0.3; 0.4), (0.4; 0.3)\}$, $P4 = \{(0.3; 0.3), (0.4; 0.4)\}$,
$P5 = \{(1.0; 0.0), (0.5; 0.5)\}$, $P6 = \{(1.0; 0.0), (0.5; 0.0)\}$

Using the similarity $s_{PE}$, it results:
$$s_{PE}(P1) = 0.80 > s_{PE}(P2) = 0.66$$
$$s_{PE}(P4) = 0.85 > s_{PE}(P3) = 0.81$$
$$s_{PE}(P6) = 0.58 > s_{PE}(P5) = 0.5$$

We discover that the pair P1 is more similar then P2, P3 is less similar then P4 and P5 is less similar then P6.
Using the similarity $s_{PH}$, it results:
$$s_{PH}(P4) = 0.85 = s_{PH}(P3) = 0.85$$
$$s_{PH}(P6) = 0.60 > s_{PH}(P5) = 0.5$$

We get that the P3 and P4 are equivalent and once again, P5 is less similar then P6.
Using the similarity $s_{PP}$, it results:
$$s_{PP}(P4) = 0.85 > s_{PP}(P3) = 0.81$$
$$s_{PP}(P6) = 0.50 = s_{PP}(P5) = 0.5$$

We find that P5 with P6 are equivalent and once again, P3 is less similar then P4. It seems that $s_{PE}$ is more credible then $s_{PH}$ or $s_{PP}$ but the formulae of $s_{PH}$ and $s_{PP}$ are simpler then $s_{PE}$ formula.

# 6 The cardinality and entropy of bipolar fuzzy set

Using the formulae of distance or similarity we can define some measures for cardinality and entropy.

## 6.1 The cardinality of bipolar fuzzy set

We will define the cardinality for a bipolar fuzzy set $A$ by:
$$n(A) = \sum_{x \in X} n(x) \quad (6.1.1)$$

where $n(x) = n(t, f, u, c)$ verifies the following conditions:

c1) $n(T) = 1$, $n(F) = 0$, $n(I) = 0.5$

c2) if $t^* > t$ then $n(t^*, 0, u, c) \geq n(t, 0, u, c)$

  if $f^* > f$ then $n(0, f^*, u, c) \leq n(0, f, u, c)$

  if $u^* > u$ then $n(t^*, f, u, 0) \leq n(t, f, u, 0)$

  if $c^* > c$ then $n(t, f, 0, c^*) \leq n(t, f, 0, c)$

c3) $n(x) = n(x^d)$ and $n(x^c) = n(x^n)$

c4) $n(x) + n(x^c) \leq 1$

c5) if $x_1 \supset x_2$ then $n(x_1) \geq n(x_2)$

where $x_1 \supset x_2$ if $\mu_1 \geq \mu_2$ and $\nu_1 \leq \nu_2$

From geometrically point of view if we denote by $\partial A$ the border between $A$ and $A^c$ we can consider as cardinality of the border the next difference:
$$n(\partial A) = n(X) - n(A) - n(A^c)$$

The following example verifies the conditions *c1, c2, c3, c4, c5*:
$$n(x) = s(x, T) \quad (6.1.2)$$

Using $s_{PE}(x, T)$, one obtain the first example:
$$n(x) = 1 - \sqrt{\left(\frac{1-t+f}{2}\right)^2 + \left(\frac{u-c}{1+u+c}\right)^2} \quad (6.1.3)$$

For fuzzy set, intuitionistic and paraconsistent fuzzy set, it results:
$$n(x) = \mu \quad (6.1.4)$$
$$n(x) = 1 - \sqrt{\left(\frac{1-\mu+\nu}{2}\right)^2 + \left(\frac{\pi}{1+\pi}\right)^2} \quad (6.1.5)$$
$$n(x) = 1 - \sqrt{\left(\frac{1-\mu+\nu}{2}\right)^2 + \left(\frac{\kappa}{1+\kappa}\right)^2} \quad (6.1.6)$$

Using $s_{PH}(x, T)$, one obtains the second example:
$$n(x) = \frac{1+t-f}{2+u+c} \quad (6.1.7)$$

For fuzzy set, intuitionistic and paraconsistent fuzzy set, it results:
$$n(x) = \mu \quad (6.1.8)$$
$$n(x) = \frac{1+\mu-\nu}{2+\pi} \quad (6.1.9)$$
$$n(x) = \frac{1+\mu-\nu}{2+\kappa} \quad (6.1.10)$$

Using $s_{PP}(x, T)$, one obtains the third example:
$$n(x) = \frac{1+t-f}{2(1+u+c)} \quad (6.1.7)$$

For fuzzy set, intuitionistic and paraconsistent fuzzy set, it results:
$$n(x) = \mu \quad (6.1.11)$$
$$n(x) = \frac{1+\mu-\nu}{2 \cdot (1+\pi)} \quad (6.1.12)$$
$$n(x) = \frac{1+\mu-\nu}{2 \cdot (1+\kappa)} \quad (6.1.13)$$

Finally, we comment the following functions used for cardinality of intuitionistic fuzzy sets:
$$n_{\min}(x) = \mu = \frac{1+t-f-u}{2} \quad (6.1.14)$$
$$n_{med}(x) = \mu + \frac{\pi}{2} = \frac{1+t-f}{2} \quad (6.1.15)$$
$$n_{\max}(x) = \mu + \pi = \frac{1+t-f+u}{2} \quad (6.1.15)$$

Only (6.1.14) and (6.1.15) verifies all the conditions of the proposed bipolar cardinality measure.

## 6.2 The entropy of bipolar fuzzy set

As well, we can define the measure for bipolar entropy. One considers the following properties for a bipolar entropy measure $e(x) = e(t, f, u, c)$:

e1) $e(T) = 0$, $e(F) = 0$,

e2) $e(I) = 1$

e3) if $t^* > t$ then $e(t^*, 0, u, c) \leq e(t, 0, u, c)$

  if $f^* > f$ then $e(0, f^*, u, c) \leq e(0, f, u, c)$

  if $u^* > u$ then $e(t, f, u^*, 0) \geq e(t, f, u, 0)$

  if $c^* > c$ then $e(t, f, 0, c^*) \geq e(t, f, 0, c)$

e4) $e(A) = e(A^c) = e(A^d) = e(A^n)$

e5) $e(U) = e(C) \geq e(I)$

The properties (*e1, e2, e3, e4, e5*) represent an extension of properties considered by De Luca and Termini for entropy of fuzzy set [4]. The bipolar entropy for a set $A$ is obtained with the subsequent formula:
$$e(A) = \frac{\sum_{x \in X} e(x)}{card(X)} \quad (6.2.1)$$

The following example verifies the conditions *e1, e2, e3, e4, e5*:
$$e(x) = 2\min(d(x, T), d(x, F)) \quad (6.2.2)$$

Using $d_{PE}(x, T)$, one obtain:
$$e(x) = \sqrt{(1-t-f)^2 + \left(\frac{2u-2c}{1+u+c}\right)^2} \quad (6.2.3)$$

We find for $U, C$ the values: $e(U) = e(C) = \sqrt{2}$. For fuzzy set, intuitionistic and paraconsistent fuzzy set, it results:
$$e(x) = 1 - |1 - 2\mu|$$

$$e(x) = \sqrt{(1-|\mu-\nu|)^2 + \left(\frac{2\cdot\pi}{1+\pi}\right)^2}$$

$$e(x) = \sqrt{(1-|\mu-\nu|)^2 + \left(\frac{2\cdot\kappa}{1+\kappa}\right)^2}$$

Using $d_{PH}(x,T)$ in (6.2.2) it results:

$$e(x) = 2\cdot\frac{1-t-f+u+c}{2+c+u} \quad (6.2.4)$$

We find for $U,C$ the values: $e(U) = e(C) = \frac{4}{3}$. For fuzzy set, intuitionistic and paraconsistent fuzzy set, it results:

$$e(x) = 1 - |1-2\mu|$$

$$e(x) = 2\cdot\frac{1-|\mu-\nu|+\pi}{2+\pi}$$

$$e(x) = 2\cdot\frac{1-|\mu-\nu|+\kappa}{2+\kappa}$$

Using $d_{PP}(x,T)$ in (6.2.2) it results:

$$e(x) = \frac{1-t-f+2\cdot u+2\cdot c}{1+c+u} \quad (6.2.5)$$

We find for $U,C$ the values: $e(U) = e(C) = \frac{3}{2}$. For fuzzy set, intuitionistic and paraconsistent fuzzy set, it results:

$$e(x) = 1 - |1-2\mu|$$

$$e(x) = \frac{1-|\mu-\nu|+2\cdot\pi}{1+\pi}$$

$$e(x) = \frac{1-|\mu-\nu|+2\cdot\kappa}{1+\kappa}$$

Using for entropy the next function:

$$e(x) = \frac{1-t-f+u+c}{1+t+f+u+c} \quad (6.2.6)$$

one gets for intuitionistic fuzzy set, the entropy defined by Szmidt and Kacprzyk [7]:

$$e_{SK}(x) = \frac{1-|\mu-\nu|+\pi}{1+|\mu-\nu|+\pi} \quad (6.2.7)$$

Using for entropy the next function:

$$e(x) = \frac{1-t-f}{1-u-c} \quad (6.2.8)$$

one gets for intuitionistic fuzzy set, the π-entropy defined by Szmidt and Kacprzyk [14]:

$$e_{SK\pi}(x) = \frac{1-|\mu-\nu|}{1-\pi} \quad (6.2.9)$$

Using for entropy the next function:

$$e(x) = u + c \quad (6.2.10)$$

one gets for intuitionistic fuzzy set, the entropy defined by Bustince and Burillo [13]:

$$e_{BB}(x) = \pi \quad (6.2.11)$$

From (6.2.6), (6.2.8) and (6.2.10) it is clear that the measures $e_{SK}$ and $e_{SK\pi}$ verifies all the conditions of bipolar entropy proposed in this paper and the measure $e_{BB}$ verifies the conditions (e1, e3, e4, e5) but it does not verify the condition (e2). Finally, we must comment the vector approach of intuitionistic fuzzy entropy proposed by Grzegorzewski and Mrowka [5]. If we take into account for entropy description the vector defined by formula:

$$e(x) = (1-t-f, u+c) \quad (6.2.12)$$

one obtains for intuitionistic fuzzy set, the following vector entropy:

$$e_{GM}(x) = (1-|\mu-\nu|, \pi) \quad (6.2.13)$$

From (6.2.12), it is observable that any norm of this vector verifies the conditions of the proposed bipolar entropy, namely (*e1, e2, e3, e4, e5*).

## 7 Conclusions

In this paper, a new distance between elements of bounded interval $[a,b]$ was constructed. Based on this distance, new measures of distance, similarity, cardinality and entropy for bipolar fuzzy sets were defined. The new measures were constructed in the framework of penta-valued representation of bipolar fuzzy sets. Also, some particular forms of cardinality and entropy were obtained for intuitionistic and paraconsistent fuzzy sets.